\documentclass[letterpaper, 10 pt, conference]{ieeeconf}  

\IEEEoverridecommandlockouts                              

\usepackage{mathtools,eqparbox}
\usepackage{multirow}

\usepackage[dvipsnames]{xcolor}
\usepackage{latexsym}
\usepackage{amssymb}
\usepackage{amsmath,mathrsfs}
\usepackage{caption}
\usepackage{epsfig}
\usepackage{bbm}
\usepackage{svg}
\usepackage{tablefootnote}

\usepackage{cite}

\usepackage{arydshln}
\usepackage{soul}
\usepackage{lipsum}

\usepackage{algorithmicx}
\usepackage[ruled]{algorithm}
\usepackage[noend]{algpseudocode}
\usepackage{blindtext}

\usepackage{verbatim}
\usepackage{hyperref}
\usepackage{graphicx}
\usepackage[lofdepth,lotdepth]{subfig}
\hypersetup{%
  breaklinks,
  bookmarksnumbered=true,
  bookmarksopen=true,
  colorlinks=true,
  linkcolor=black,
  urlcolor=black,
  citecolor=black
}
\usepackage{subfiles}

\usepackage{arydshln}

\newcommand{\algmargin}{\the\ALG@thistlm}
\algnewcommand{\parState}[1]{\State%
    \parbox[t]{\dimexpr\linewidth-\algmargin}{\strut\hangindent=\algorithmicindent \hangafter=1 #1\strut}}
\newcommand\blfootnote[1]{%
  \begingroup
  \renewcommand\thefootnote{}\footnote{#1}%
  \addtocounter{footnote}{-1}%
  \endgroup
}

\begin{document}
\bstctlcite{IEEEexample:BSTcontrol}
%

\title{\textbf{PIAug} - \textbf{P}hysics \textbf{I}nformed \textbf{Aug}mentation for Learning Vehicle Dynamics for Off-Road Navigation}

\author{Parv Maheshwari$^{1,2}$, Wenshan Wang$^{2}$, Samuel Triest$^{2}$, Matthew Sivaprakasam$^{2}$,\\ Shubhra Aich$^{2}$, John G. Rogers III$^{3}$, Jason M. Gregory$^{3}$ and Sebastian Scherer$^{2}$ 
\thanks{* This work was supported by ARL awards \#W911NF1820218 and \#W911NF20S0005.}%
\thanks{$^{1}$ Department of Mathematics, Indian Institute of Technology Kharagpur. parvmaheshwari2002@iitkgp.ac.in}%
\thanks{$^{2}$ Robotics Institute, Carnegie Mellon University, Pittsburgh, PA, USA. \{parvm,wenshanw,striest,msivapra,saich,basti\}@andrew.cmu.edu}%
\thanks{$^{3}$ DEVCOM Army Research Laboratory, Adelphi, MD, USA. \{john.g.rogers59.civ, jason.m.gregory1.civ\}@army.mil}%
}

\maketitle
\IEEEpeerreviewmaketitle

\begin{abstract}
\blfootnote{Accompanying Video: \href{https://youtu.be/r0Dq61DIWZQ}{\textcolor{blue}{https://youtu.be/r0Dq61DIWZQ}}}
Modeling the precise dynamics of off-road vehicles is a complex yet essential task due to the challenging terrain they encounter and the need for optimal performance and safety. Recently, there has been a focus on integrating nominal physics-based models alongside data-driven neural networks using Physics Informed Neural Networks. These approaches often assume the availability of a well-distributed dataset; however, this assumption may not hold due to regions in the physical distribution that are hard to collect, such as high-speed motions and rare terrains. Therefore, we introduce a physics-informed data augmentation methodology called PIAug. We show an example use case of the same by modeling high-speed and aggressive motion predictions, given a dataset with only low-speed data. During the training phase, we leverage the nominal model for generating target domain (medium and high velocity) data using the available source data (low velocity). Subsequently, we employ a physics-inspired loss function with this augmented dataset to incorporate prior knowledge of physics into the neural network. Our methodology results in up to 67\% less mean error in trajectory prediction in comparison to a standalone nominal model, especially during aggressive maneuvers at speeds outside the training domain. In real-life navigation experiments, our model succeeds in 4x tighter waypoint tracking constraints than the Kinematic Bicycle Model (KBM) at out-of-domain velocities.
\end{abstract}

\section{Introduction}
Autonomous off-road driving has been a field of great interest \cite{triest2022tartandrive, TerrainNet,mutimodal_model, Stanley,off_road_static,off_road_world_model, HDIF} due to applications in various industries, like
agriculture \cite{agri_1,agri_2,agri_3}, search and rescue \cite{rescue_1,rescue_2}, and military operations \cite{military_1,military_2,military_3}. It frequently involves performing challenging maneuvers, like high-speed travel over rough terrain and sharp turns at high yaw rates, where the dynamics of a robot, if not accurately modeled, can lead to significant errors and potential fatal accidents.

Traditional vehicle models enable robots to navigate and interact with their surroundings by incorporating general physics equations based on mechanics and dynamics. Hence, they do not overfit on any particular distribution but rather generalize equivalently across domains. However, their fundamental modeling assumptions can impose limitations in challenging scenarios. For example, the widely used physics-based vehicle model, the Kinematic Bicycle Model (KBM), assumes negligible tire slip. This assumption gets violated during sharp turns and on low friction areas, resulting in large prediction errors. These models can only use low-dimensional modalities like odometry, wheel encoders, and IMU data.  
\begin{figure}[t]
\captionsetup[subfigure]{}
\centering
\subfloat[][Moving on high pitched incline\\ with $V_{S}=4.7$m/s \textbf{,} $V_M=3.5$m/s]{
\includegraphics[width=0.23\textwidth]{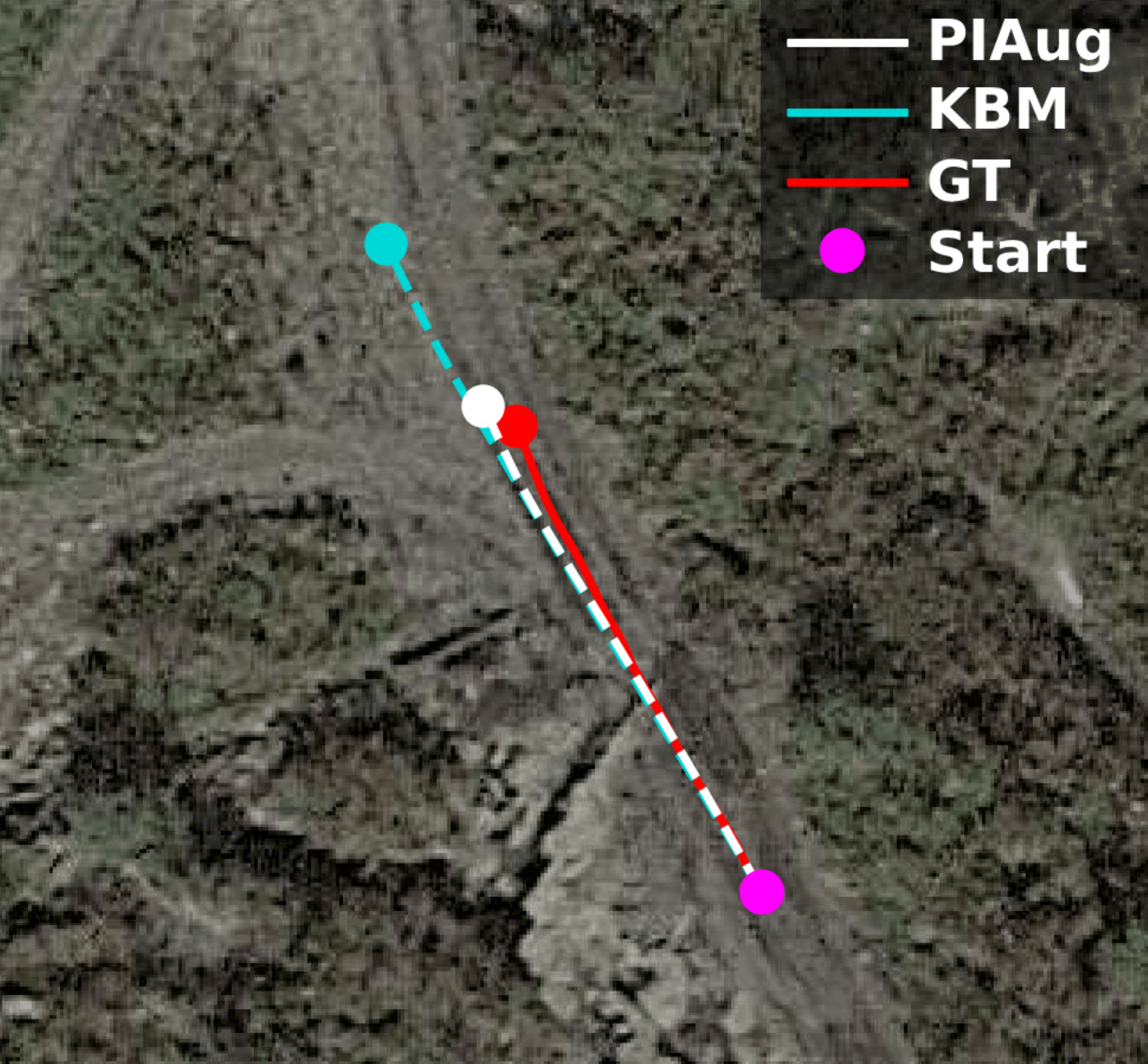}}
\subfloat[][Taking a turn aggressively with \\\phantom{aaa} $V_S=6.59$m/s \textbf{,} $V_M=5.42$m/s]{
\includegraphics[width=0.228\textwidth]{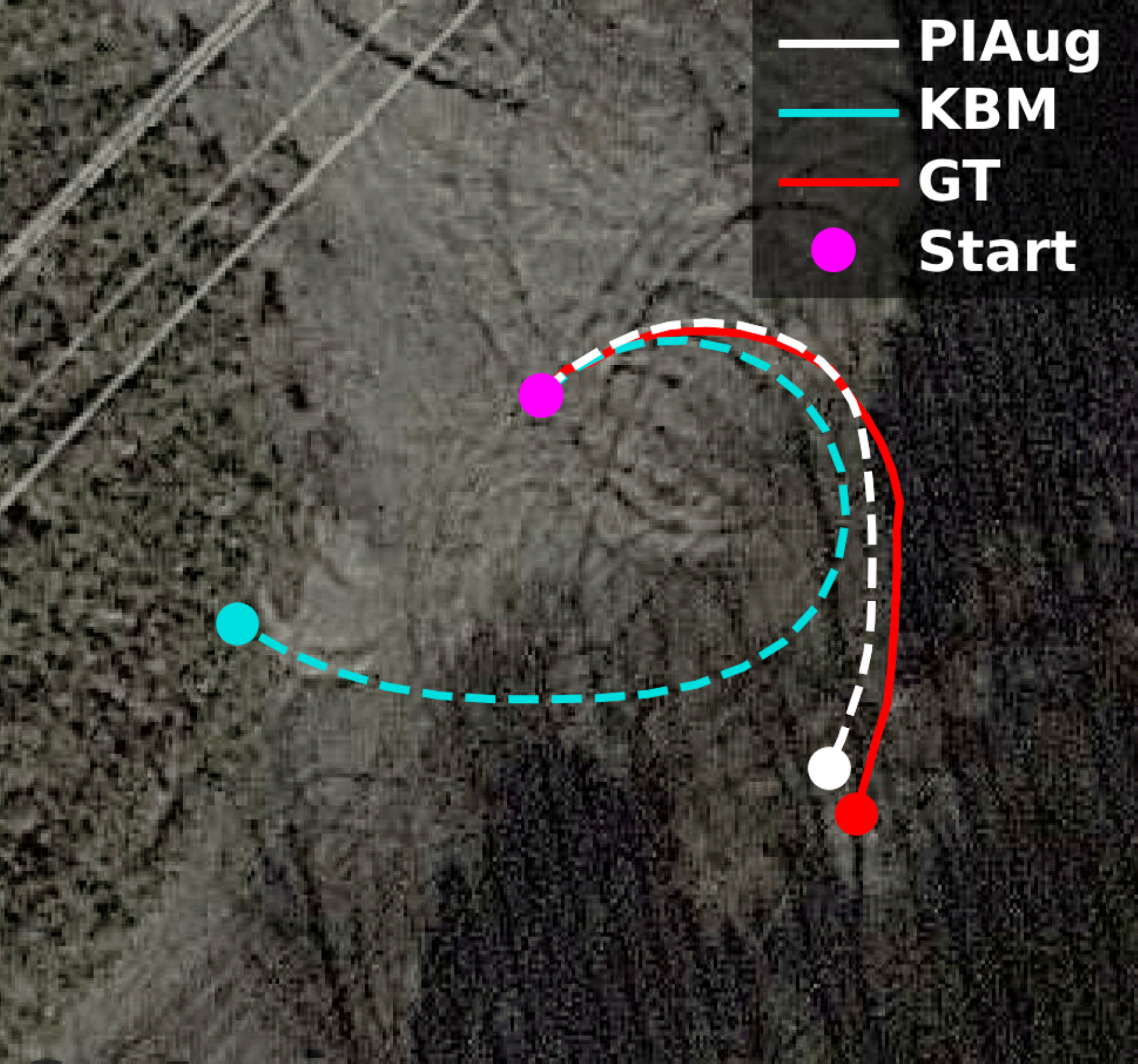}
}
\caption{\textit{Trajectories predicted over 5s horizon by PIAug and KBM with respect to the ground truth (GT). Here, $V_S$ and $V_M$ denote the starting and average speed of ground truth. In both cases, PIAug shows robustness to out-of-domain scenarios with respect to speed since we only use low-speed ($<=3$m/s) data to train PIAug.}}
\vspace{-1.2em}
\label{fig: front_page}
\end{figure}
As an alternative to physics-based models, \cite{triest2022tartandrive, MultimodalVM} have explored pure data-driven neural networks to bypass the assumptions and limitations of traditional models, possibly by leveraging multiple low and high-dimensional data like terrain maps. These neural network (NN) methods precisely capture the complex vehicle dynamics on in-domain samples but experience a rapid decline in prediction accuracy as gap widens between validation samples and training distribution. 

To gain benefits of both neural networks and expert physics knowledge, recent works use Physics Informed Neural Network (PINN) \cite{PINN_og} to model system dynamics of UAVs and soft robots \cite{PI-TCN, RAMPNet,PI_soft_robot}. By introducing ``physics-inspired" loss, in surplus to standard task-specific loss, PINNs enhance the models' applicability while maintaining their accuracy within their specific domain. However, we find that this improvement in cross-domain generalization is limited by the dataset distribution used to determine the loss.

In off-road driving, it is hard to collect balanced data since driving at high speeds and aggressive turning is risky and tedious. In such cases, where access to certain distributions is not possible, augmenting the dataset can help achieve a more uniform distribution. 
In this work, exemplifying velocity distribution as different domains, we answer the question - ``Can a vehicle model trained on low-velocity data (source domain) excel in 
medium and 
higher-velocity (target domain) maneuvers?".

To generate a viable ``target dataset," we need an alternative approach to standard synthetic generation techniques, i.e., random perturbations over training data and using the simulator to generate a balanced dataset. However, our task to transfer data from the low-velocity domain to a higher-velocity domain would require adding large perturbations to the state and action sequence, which cannot guarantee a kino-dynamically feasible multi-step sequence.
Moreover, using simulators becomes difficult with larger-scale robots due to the challenge of creating accurate vehicle simulation models.

As an alternative to these approaches, \cite{PhyAug} demonstrates leveraging a physics prior for generating target domain data from the training data to counter domain shifts. However, they generate this ``augmented data" before the training stage; hence, the knowledge of physics prior is never explicitly added to the NN. Taking inspiration from both PINNs and \cite{PhyAug}, we propose \textit{PIAug - Physics Informed Augmentation} to generate a target dataset while extracting the complete potential of the physics prior and NNs. For this, we use a physics model to generate an augmented dataset during training. Unlike \cite{PhyAug}, this allows us to calculate a physics-inspired loss similar to PINN, which can then be backpropagated.

The main contributions of this work are as follows:
\begin{itemize}
    \item To our knowledge, we designed the first-ever multi-modal PINN using low and high-dimensional inputs to learn the system dynamics of our off-road robot. 
    \item We introduce a data augmentation method, PIAug - a physics model-based augmentation of the dataset during training, resulting in a physics-informed NN trained over out-of-distribution data.
    \item We compare the predictive performance of PIAug against several models. We also evaluate PIAug on full-scale navigation experiments and showcase its benefits over KBM in performing aggressive maneuvers even at out-of-domain velocities (Fig. \ref{fig: front_page}). 
\end{itemize}

\section{Related Work}
\label{related}
Broadly, all the works for modeling system dynamics can be classified as - physics-based, data-driven, or a hybrid approach combining data-driven and physics models.

\subsection{Physics-Based Vehicle Models}
KBM and its variants are the prevailing models in this category, but to overcome their simplified kinematic assumptions, recent works \cite{phy_vm_1,phy_vm_2,phy_vm_3,phy_vm_4} have explored high-fidelity models. However, a significant drawback of these models is their need for comprehensive data on the robot's mechanical geometry and actuator limitations. Obtaining these can be challenging in some cases. Jeon et al.\cite{phy_vm_5} present a framework that enhances tire force estimation by using onboard navigation sensors to employ the Magic Formula\cite{magic_formula}. However, this method requires wheel encoders, which limits its applicability to robots with these sensors. Also, highly nonlinear equations in such models result in slower run times as compared to KBM.

 
\subsection{Data-Driven Neural Networks}

Wang et al. \cite{sean_data} uses a data-driven NN for terrain traversal without perceptual information. Triest et al. \cite{triest2022tartandrive} explored the role of additional modalities like First Person View (FPV) Image, top-down view RGB and height maps, and time series data like IMU in improving prediction accuracy. Our neural network architecture is similar to [2], with some modifications explained in Section \ref{data_vm}.
 
\subsection{Hybrid Models}
Similar to the approach in \cite{korea_plain}, there are works that utilize neural network outputs to predict inputs to a final Kino-dynamic layer and, therefore, are also bound by the assumptions of this layer. For instance, in \cite{korea_plain}, the last layer does not account for changes in the robot's altitude, roll, and pitch, along with their impact on other state variables, and hence the robot's motion is constrained to a 2D plane. 
Some works like \cite{korea_offroad} leverage a roll-pitch-yaw physics model along with terrain information to overcome the assumptions of a 2D final layer. A point of difference between our works is \cite{korea_offroad} only showcases their method over simulated data and cannot consider other modalities like RGB maps and FPV images or terrain maps.


An alternative approach to add physics knowledge to NNs is to use residual methods. Here, NN is employed to capture and model the uncertainties that exist between the predictions of the physics model and the ground truth. For instance, \cite{knodempc} demonstrates the application of this approach in the context of Unmanned Aerial Vehicles (UAVs). 

More recent works \cite{PI-TCN} have shown better performance of PINNs over residual methods. Here, physics models are used to add explainability to NNs by adding a physics-inspired loss during training in addition to the standard data-driven loss. Another key distinction from residual methods is that PINNs predict absolute future states during inference instead of serving as a residual term to the physics model. While \cite{ModalPINN} uses a single high-dimensional modality as input to reconstruct a fluid's flow pattern, we can use of multiple high-dimensional modalities as inputs to a PINN.

\section{Methodology}
\label{method}
\subsection{Extended Kinematic Bicycle Model}
 We employ an extended KBM (Eq. \ref{general KBM}), which expands the typical state representation $[x_i, y_i, \psi_i]$ with additional variables $[v_i, \delta_i]$. $x_i$ and $y_i$ denote global positional coordinates, $\psi_i$ and $v_i$ represent the yaw and speed of the vehicle in the global frame, and $\delta$ denotes the steering angle of the wheels at timestep $i$. 
 These supplementary state variables enable us to incorporate $[T_i,\delta_i^{target}]$ as actions ($U_i$), where $T_i$ represents throttle and $\delta_i^{target}$ denotes the target steering angle. Since throttle is a lower-level control, it is chosen as the desired action instead of a target velocity as in a standard KBM. While calculating the rate of change of speed ($\dot{v_i}$), we consider the effect of throttle, a combined velocity proportional effect of engine braking and drag force, along with friction, gravity ($g$) and pitch ($\theta_i$). These effects are scaled appropriately using constants ($K_t, K_b, K_f, K_g$).
 Our KBM also models steering wheel actuation, using $K_s$, and eliminates the assumption that the target steering is achieved instantaneously as in a standard KBM, which directly uses $\delta_i^{target}$ instead of $\delta_i$ to predict yaw rate. 
\begin{figure*}[h]
\centering
\includegraphics[width=0.85\textwidth]{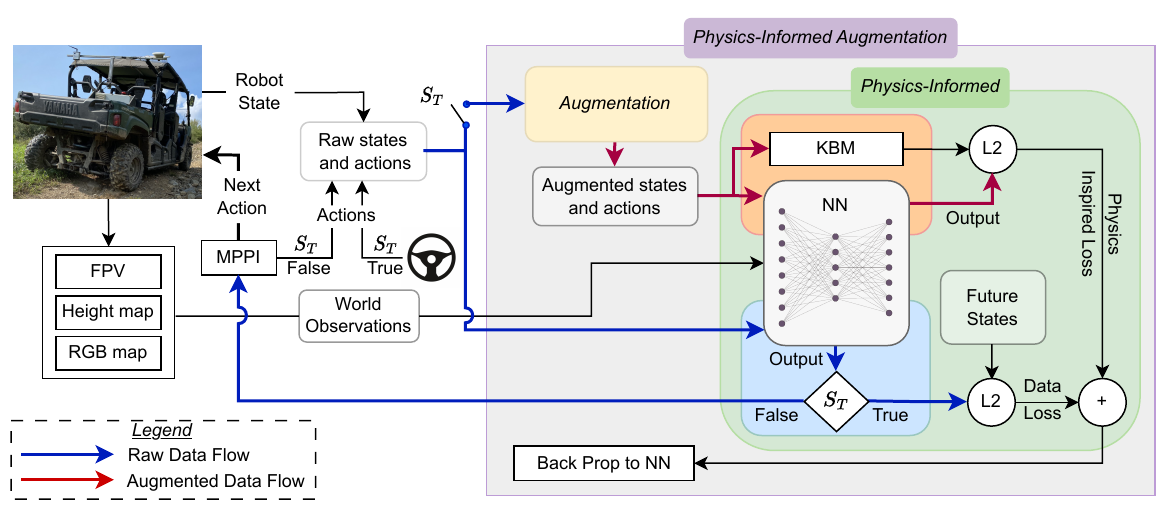}
\caption{\textit{ An overview of our methodology. During training ($S_T$ = True), the raw states and actions (from a dataset) are fed to an augmentation methodology to generate an augmented state and sequence of actions. To generate the ground truth label of this augmented data, KBM (a physics model) is used. For both raw and augmented data sequences, a shared NN, along with common observations, are used to predict future states. We linearly combine the data-driven loss with the physics-informed loss. For clarity of the figure, we have not connected the backpropagation step back to the NN. At test time ($S_T = False$), there is no augmentation, and MPPI generates the raw actions. The predicted ground truth for the raw data is then sent back to MPPI to evaluate their costs.
}}
\vspace{-0.8em}
\label{fig: flowchart}
\end{figure*}
 \begin{subequations}
 \label{general KBM}
 \begin{flalign}
    f(X_i,U_i) =&\  \dot{X_i}\ =\ \begin{bmatrix}v_i*cos\psi_i\\v_i*sin\psi_i\\(v_i\tan\delta_i)/L\\\dot{v_i}\\\dot{\delta_i}\end{bmatrix} \\
    \dot{v_i} =&\ K_t * T_i - K_b * v_i - K_f * sign(v_i) * \cos(\theta_i) \nonumber
    \\ & \hphantom{K_t * T_i - K_b * v_i \ }- K_g * g* \sin(\theta_i) \\
    \dot{\delta_i} =&\ K_s * (\delta^{target}_i - \delta_i)
\end{flalign}
\end{subequations}

\subsection{Data driven Vehicle Model}
\label{data_vm}

Building upon the ideas presented in \cite{triest2022tartandrive}, we employ a NN with multimodal inputs to estimate vehicle dynamics. One of the notable differences in our model is the incorporation of all components of linear $\textbf{v}$ and angular $\textbf{w}$ velocities, 
in addition to the current steering angle ($\delta$), into the state space. This choice was motivated by KBM, as it uses velocity and current steering angle to predict the future position and orientation of the robot. Further, we have chosen a recently introduced 6-state representation of rotation, denoted as $\textbf{r}$, as proposed in \cite{continuous_rotation}. 
This is a continuous representation, unlike quaternions ($\textbf{q}$), which possesses discontinuity. This choice is made to take advantage of the neural network's ability to perform better when dealing with continuous data over discontinuous data. Consequently, the new state vector for our neural networks consists of sixteen elements: 
[$\textbf{p},\textbf{r},\textbf{v},\textbf{w},\delta$] compared to \cite{triest2022tartandrive} which utilizes a seven-state vector: [$\textbf{p},\textbf{q}$]. 
This enhanced state allowed us to achieve optimal predictive accuracy even without time series data such as wheel encoders and IMU. Additionally, we enhance our model by employing a 4-channel height map that encompasses [min, max, mean, std] values, in contrast to \cite{triest2022tartandrive}, which utilizes only a 2-channel height map consisting of [min, max].

\subsection{Physics-Informed Vehicle Model}
We formulate the Physics-Informed vehicle model as a combination of physics-inspired and data-driven loss functions (Eq \ref{pinn_loss}). In our context, $B_D$ represents a real-world dataset while $B_{P}$ can either represent a real-world dataset for a PINN model or an augmented dataset for a PIAug model (Section \ref{PIAug}). We use $B_P$ and $B_D$, for determining physics-informed ($L_{P}$) and data-driven ($L_{D}$) losses, respectively. Here, $\phi(X_{k},O_{k},\text{\boldmath{$U_{k}$}})$ represents the data-driven NN, and the physics model is signified by $F(X_{k},\text{\boldmath{$U_{k}$}})$ and $f$ (Eq. \ref{general KBM}). $k$ and $T$ denote the current timestep and prediction horizon, respectively. $\lambda_{PI}$ is the weight of $L_P$, in its linear combination($L$) with $L_D$. For better accuracy, we use the Midpoint method\cite{midpoint}
to evaluate the integral in Eq. \ref{pinn_loss}b.

A data sequence is defined as $\{X_k,O_k,\text{\boldmath{$U_k$}},\text{\boldmath{$Y_k$}}\}$ where $X_k$ and $O_k$ are the current state and observations defined as all the other multiple modalities required by the NN. In our case, this comprises \{FPV Image, RGB map, Height map\} similar to \cite{triest2022tartandrive}. {\boldmath{$U_k$}} $= [U_k,U_{k+1}...,U_{k+T-1}]$ are the $T$ actions, while  {\boldmath{$Y_k$}} $= [X_{k+1},X_{k+2}...,X_{k+T}]$ denotes ground truth for the future trajectory. The predicted states of the data sequence using a model, $m$, are represented as {\boldmath{$\hat{X}_k^m$}} $=[\hat{X}_{k+1}^m,\hat{X}_{k+2}^m,...,\hat{X}_{k+T}^m]$, which can be calculated as shown in Eq. \ref{pinn_loss}. An important note here is that since the state space of the NN and the physics model are different, we convert the prediction of the NN to KBM's state space in Eq. \ref{pinn_loss}c. This is represented by {\boldmath{$(\hat{X}_i^{NN})_{KBM}$}}.
\begin{subequations}
\label{pinn_loss}
\begin{flalign}
    \text{\boldmath{$\hat{X}_{k}^{NN}$}}\ &=\ \phi(X_{k},O_{k},\text{\boldmath{$U_{k}$}})
    \\\text{\boldmath{$\hat{X}_{k}^{KBM}$}} &=\  F (X_k,\text{\boldmath{$U_k$}})
    \\ where &\ \hat{X}_{k+l+1}^{KBM}\ =\ \hat{X}_{k+l}^{KBM}\ + \int_{t=0}^{t=1}f (X_{k+l},U_{k+l})\ dt , \nonumber
    \\&\ \hat{X}_{k}^{KBM}\ = \ X_k, \text{ and } l \in [0,1...,T-1]\nonumber
    \\ L_P &= \frac{1}{|B_{P}|}\sum_{i=1}^{|B_{P}|}||\text{\boldmath{$(\hat{X}_i^{NN})_{KBM}$}} 
     - \text{\boldmath{$\hat{X}_{i}^{KBM}$}}||^2
    \\ L_D &= \frac{1}{|B_{D}|}\sum_{j=1}^{|B_{D}|}||(\text{\boldmath{$\hat{X}_{j}^{NN}$}} -\text{\boldmath{$Y_j$}})||^2
    \\ L &= L_D + \lambda_{PI} L_P
\end{flalign}
\end{subequations}

\subsection{Physics-Informed Data Augmentation}
\label{PIAug}
To improve NN's generalization, the physics model can be used to generate multi-step ground truth predictions for out-of-distribution scenarios. This allows the model to learn from the physics model, akin to PINN, while also augmenting the training dataset, allowing it to be robust to domain shift. We illustrate domain shifts in predictive and navigational performance across different mean speed groups.

 For this, we will replace a real-world dataset with an augmented dataset to use as $B_{P}$. Fig. \ref{fig: flowchart} provides an overview of our methodology. The physics model requires an initial state, initial observations,  and a set of actions to achieve this. Taking inspiration from our physics model, it is known that speed and yaw rate can be approximated as linearly proportional. Therefore, to generate the desired initial state in the augmentation module (Yellow box in Fig. \ref{fig: flowchart}), we apply a transformation over the initial state of the mini-batch ($X_{RAW}$) sampled from the training dataset (Here we drop the timestep subscript, $k$ for a cleaner representation). This transformation is a random scaling sampled from a uniform distribution: $U(2.5,4)$. The scaling is kept standard for all components of linear and angular velocity. The action sequences {(\boldmath{$U_{AUG}$})} used in evaluating this physics loss, however, are the same as those of the training batch {(\boldmath{$U_{RAW}$})}. A note here is that {\boldmath{$U_{AUG}$}} can easily be generated using the Gaussian or Ornstein-Uhlenbeck process \cite{uhlenbeck1930theory} if it is desired that the actions follow a preferred distribution.

\begin{figure}[h!]
\centering
\includegraphics[width=0.4\textwidth]{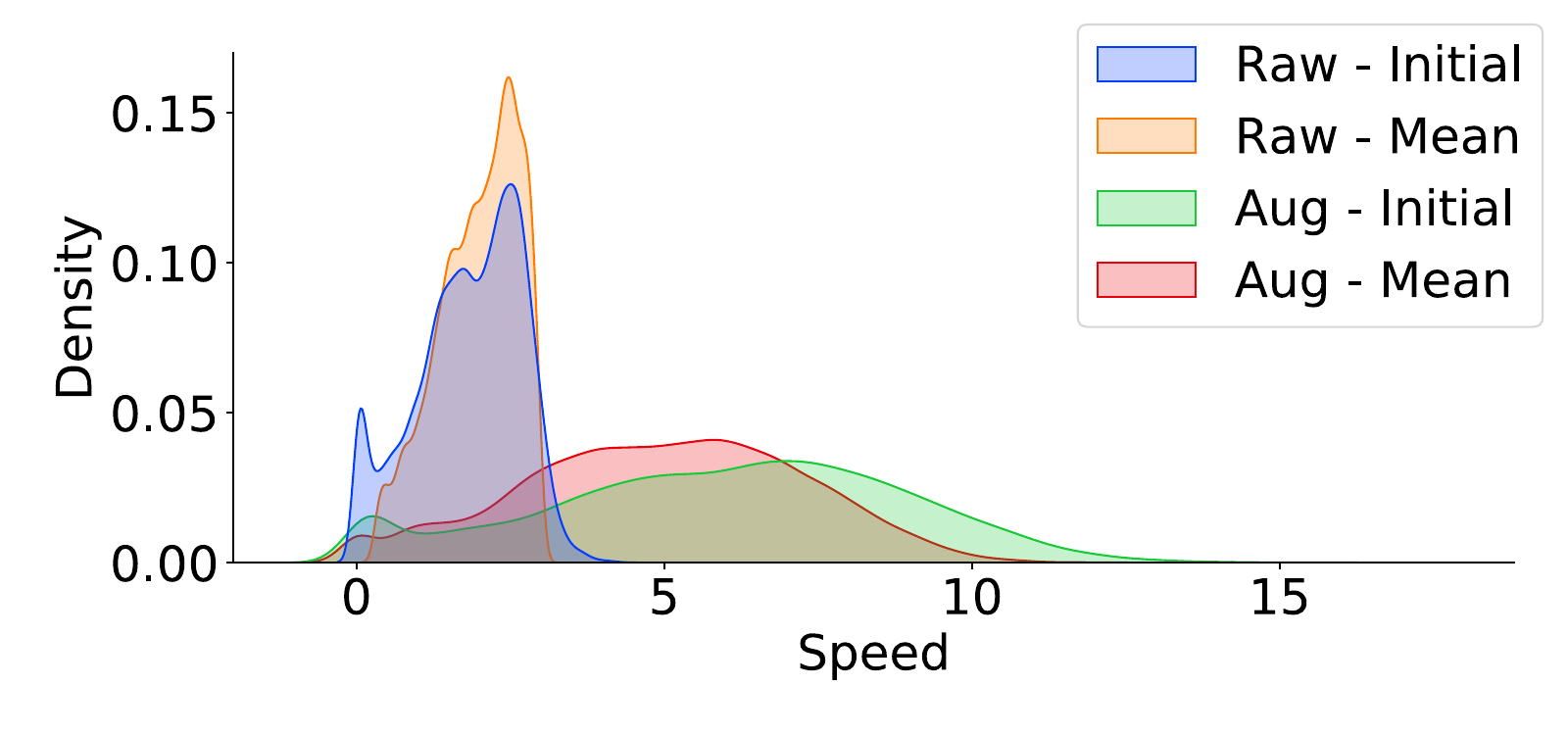}
\vspace{-0.4cm}
\caption{\textit{The density ($\rho$) distribution of initial and mean speeds.``Raw" denotes the training dataset, which exclusively contains low-velocity data. ``Aug" represents the augmented dataset created using KBM with a velocity-scaled initial state. It possesses a balanced distribution of initial and mean speeds in the trajectories over the imbalanced Raw dataset. Since there 
exists some data with a negative initial velocity ($\rho(Raw_{Initial} <0) \neq 0$), augmentation can also result in a negative mean speed ($\rho(Aug_{Mean} <0) \neq 0$). 
}}
\vspace{-0.4 cm}
\label{fig: colloc_speed}
\end{figure}

The augmented initial state ($X_{AUG}$), combined with the original observation ($O_{RAW}$) and actions {(\boldmath{$U_{AUG}$})} can then be used to predict the future states ($Y_{AUG}$) similar to {\boldmath{$\hat{X}_k^{KBM}$}} in Eq. \ref{pinn_loss}b. Fig. \ref{fig: colloc_speed} shows the distribution of initial and mean velocity for raw($B_D$) and augmented($B_P$) datasets where we use {\boldmath{$Y_{RAW}$}} and {\boldmath{$Y_{AUG}$}} to calculate mean velocities of the respective data.
\section{Hardware Implementation}
\label{implementation}
We evaluated the performance of our models on a Yamaha Viking ATV modified for autonomous driving by Mai et al. \cite{Mai-2020}. We use a Velodyne UltraPuck LiDAR sensor for collecting point clouds, a Multisense S21 camera system for FPV Stereo and RGB images, and a NovAtel PROPAK-V3-RT2i GNSS unit for global positioning. Our navigation stack ran on a 12th-gen Intel i7 CPU and an NVIDIA 3080 laptop GPU. The navigation stack can be broken down into 
\subsubsection{Costmapping}
We use Super Odometry \cite{zhao2021super} to align unregistered lidar scans in the vehicle's initial frame. These scans are then converted into gridmaps and further processed into costmaps based on a predefined lethal height threshold.
\subsubsection{RGB and Height Maps}
\label{vo_section}
Similar to \cite{triest2022tartandrive}, we use TartanVO \cite{wang2020tartanvo} to create top-down RGB maps and height maps, using RGB and Stereo FPV images. These maps serve as inputs for our neural network-based vehicle models.
\subsubsection{Control}
Much like the approach described in \cite{sam_irl}, we adapt MPPI (Model Predictive Path Integral) from Williams et al. \cite{williams2017information} for navigation. Our modified MPPI incorporates multiple waypoints to improve cost-to-goal computation and makes effective use of our long prediction horizon.
\section{Experiments, Results, and Analysis}
\label{experiment}
\subsection{Preliminaries}
\subsubsection{Dataset}
\label{dataset}
We used a subset of TartanDrive\cite{triest2022tartandrive} comprising only of the low-velocity trajectories as our training dataset. An additional dataset was gathered with a balanced velocity distribution to assess the capability of the model for both in and out-of-distribution scenarios.

\subsubsection{Errors}
All the reported prediction errors are the averages of errors within the multi-step horizon.
For a fair comparison between KBM and the NN-based vehicle models, these errors encompass various aspects, including position ($\Delta p$), yaw ($\Delta \psi$), and velocity ($\Delta v$). $\Delta p$ quantifies the $L_2$ norm of prediction errors of both the $\Delta x$ and $\Delta y$. We do not report steer angle error as an error of magnitude of the order of $10^{-2}$rad is constantly exhibited.
\begin{table}[t]
    \centering
    \vspace{0.3cm}
    \begin{tabular}{c|c||c|c|c}
        Attribute& Symbol& Low& Med & High \\
        \hline &&&&\\[-1.5ex]
         $|v|$ : Velocity&V &[0,3]& (3,5]& (5,7]\vspace{0.2em}\\
         $|\theta|$ : Pitch&$\Theta$ &[0,0.05]& (0.05,0.12]&(0.12,1.57]
         \vspace{0.2em}\\
         $|\dot\psi|$ : Yaw Rate&$\Psi$ &[0,0.05]&(0.05,0.12] &(0.12,1.57]
    \end{tabular}
    \caption{\textit{Thresholds for categorizing various attributes of a sequence into low, medium, or high groups. 
    }}
    \label{tab: groups}
\end{table}
\begin{figure}[t]
\captionsetup[subfigure]{labelformat=empty}
\centering
\vspace{-0.5em}
\subfloat[][]{
\includegraphics[width=0.43\textwidth]{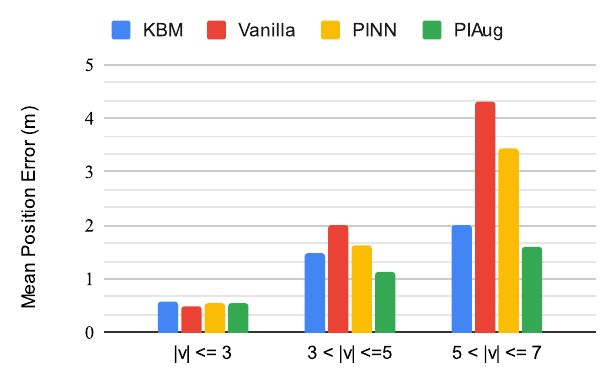}
}
\vspace{-14em}
\subfloat[][]{
\includegraphics[width=0.43\textwidth]{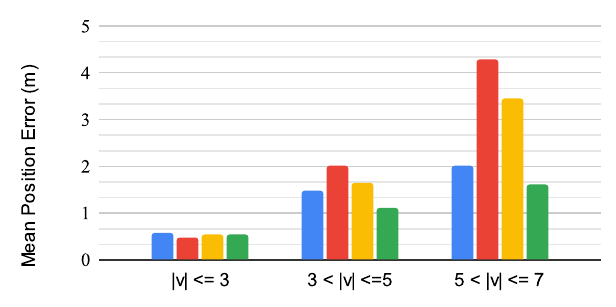}
}
\vspace{-3em}
\subfloat[Subfigure 2 list of figures text][]{
\includegraphics[width=0.43\textwidth]{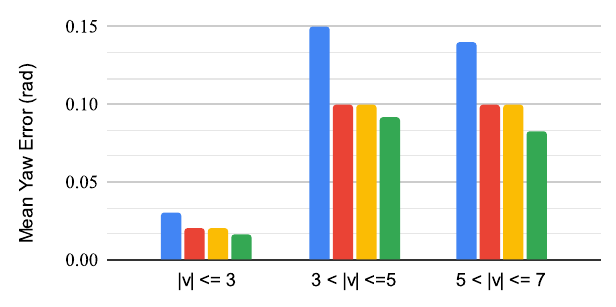}
}
\vspace{-3em}
\subfloat[Subfigure 3 list of figures text][]{
\includegraphics[width=0.43\textwidth]{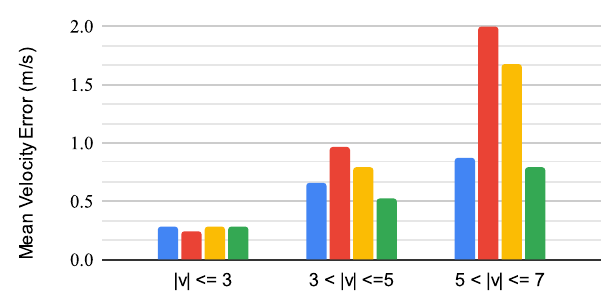}
}
\vspace{-1em}
\caption{\textit{Comparison of models trained on $<$3m/s trajectories. PIAug performs better than all other models in out-of-distribution trajectories.}}
\vspace{-1.em}
\label{fig: augmentation}
\end{figure}
\subsection{Prediction Analysis over Different Velocity Domains}
\label{open_loop_analysis}
\begin{figure}[t]
\centering
\vspace{0.5cm}
\includegraphics[width=0.48\textwidth]{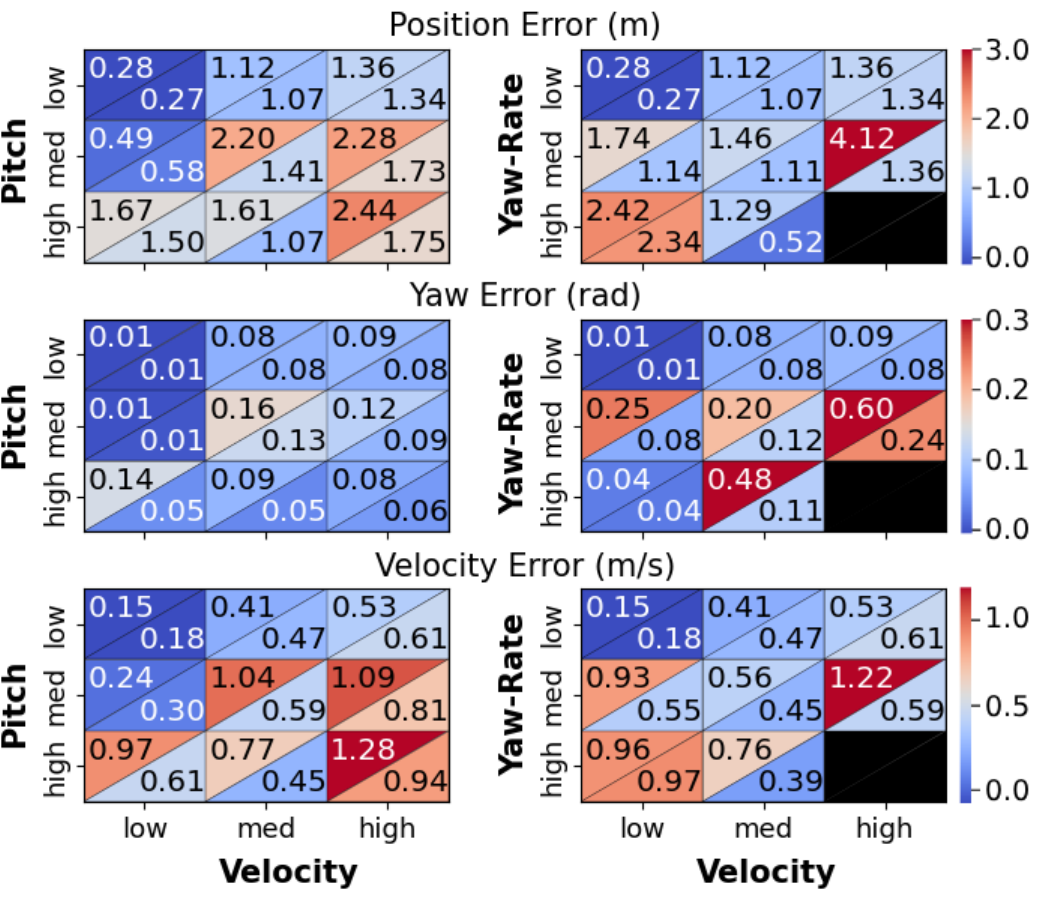}
\caption{\textit{Prediction errors in position (top), yaw (middle), and velocity (bottom) for KBM (upper half) and PIAug (lower half). Left:  Navigating straight ($\Psi_{low}$) at varying speeds over slopes with different inclines. Right: Turning at different speeds and yaw rates on flat terrain ($\Theta_{low}$). The black mask over $( V_{high},\Psi_{high},\Theta_{low})$ on the right represents no data collection to preserve driver safety. The most significant improvements (up to 67\% achieved in case of position error of $(V_{high},\Psi_{med})$ on the right) are near the bottom right half on each metric of both columns, which corresponds to the most difficult scenarios like driving fast on rugged terrains (left) and aggressive turning (right). 
}} 
\label{fig: heatmap}
\end{figure}

 As mentioned in Section \ref{PIAug}, we will divide data sequences into domains majorly on the basis of the mean speed of their ground truth $\large(|v|\large)$. Table \ref{tab: groups} lists the different domains based on velocity. We evaluate the performance of various models in Fig. \ref{fig: augmentation}, in which the vanilla model denotes a data-driven variant of PINN where we use $\lambda_{PI} = 0$ during training (Eq. \ref{pinn_loss}e). We can infer that even though PINN is better than its vanilla variant, the performance improvement is not as substantial as compared to the case of PIAug. This vast improvement between PINN and PIAug can be directly attributed to our method of scaling the initial state to other domains and hence also emphasizes the necessity to evaluate the physics-inspired loss on a well-distributed dataset. Apart from this, Fig. \ref{fig: augmentation} also shows how the performance of Vanilla NN degrades quickly as the domain shift increases. We omit Vanilla NN and PINN from subsequent analyses due to these large errors.
 

\subsection{Comparisons of vehicle models on different maneuvers}
\label{kbm_scenarios}
To highlight the complexity of different scenarios, apart from distributing them only on the basis of velocity as done in Section \ref{open_loop_analysis}, we categorize each sequence into more subgroups. This is done on the basis of combinations of the average of magnitudes of velocity $\large(|v|\large)$, pitch$\large(|\theta|\large)$, and yaw rate $\large(|\dot\psi|\large)$. The threshold values of these groups are defined in Table \ref{tab: groups}. This results in 27 subgroups denoted as $\{(V_{p},\Theta_{q},\Psi_{r}) : p,q,r \in \{low, med, high\}\}$. Fig. \ref{fig: heatmap} compares the prediction error of KBM and PIAug over two subsets of these groups (($V_p,\Theta_q,\Psi_{low}$) and ($V_p,\Theta_{low},\Psi_r$)). The first rows of the left/right heatmaps in Fig. \ref{fig: heatmap} are the same since these rows represent the scenarios $\in (V_{low},\Theta_{low},\dot\Psi_{low})$.

We can see that KBM exhibits significant prediction errors in scenarios containing high difficulty of any attribute due to its reliance on kinematic assumptions. In contrast, PIAug demonstrates considerably higher prediction accuracy (up to 67\%) even on subgroups containing $V_{high}$. This is intriguing as PIAug was trained exclusively on data from $V_{low}$. Even beyond these scenarios, PIAug consistently outperforms KBM in most groups, with comparable performance in the remaining cases. 
\subsection{Real-Time Frequency Comparison}
NN-based vehicle models like PIAug enhance predictions and, if optimized, reduce inference time compared to KBM. This accelerates the navigation stack, improving performance and enabling more samples in MPPI for better control solutions.
Table \ref{tab: freq} compares inference times for KBM and a NN-based vehicle model during MPPI trajectory rollouts. NN-based model rollouts, implemented with GPU-based RNNs\cite{PyTorch}, can offer a speed advantage over solving integrals in the case of KBMs if they are well-optimized. In MPPI, since all samples share an initial state, it allows encoding a single sample during inference and using the result with all sampled actions. This approach (denoted as NN\_OPT) significantly reduces the required GPU memory and inference time. To optimize control solutions with accurate PIAug, referencing Table \ref{tab: freq}, we select 2048 MPPI rollout samples. This choice mitigates GPU memory challenges tied to PIAug's accuracy, which could occur with 4096 samples.
\begin{table}[t]
    \centering
    \vspace{0.2cm}
    \begin{tabular}{c||cc|cc|cc}
        &\multicolumn{6}{c}{Number of MPPI Samples}\\
        \hline &&&&&&\\[-1.5ex]
        Model Type&\multicolumn{2}{c|}{1024}&\multicolumn{2}{c|}{2048}&\multicolumn{2}{c}{4096}\\
        &&&&&&\\[-1.5ex]
        &Time\tablefootnote{Inference time (in seconds) during the MPPI rollout stage.}&Mem\tablefootnote{Amount of GPU memory required (in GB)}&Time&Mem&Time&Mem\\
        \hline&&&&&&\\[-1.5ex]
        KBM\_CPU & 0.058 & 0.00 & 0.066 & 0.00 & 0.132 & 0.00  \\
        KBM\_GPU & 0.074 & 0.598 & 0.072 & 0.622 & 0.069 & 0.622  \\
        NN & 0.166 & 3.806 & N/A & N/A & N/A & N/A  \\
        NN\_OPT & \textbf{0.027} & 2.064 & \textbf{0.035} & 2.510 & \textbf{0.052} & 4.996  \\
    \end{tabular}
    \caption{\textit{NN-based models are more precise than KBM, and an optimized implementation provides 2x faster inference time.
    N/A represents insufficient ($>$6GB) GPU memory.}}
    \vspace{-0.25cm}
    \label{tab: freq}
\end{table}
\subsection{Hardware Experiments}
\label{navigation_experiments}
\begin{figure}[t]
    \centering
    \vspace{0.5em}
    \includegraphics[width = 0.4\textwidth]{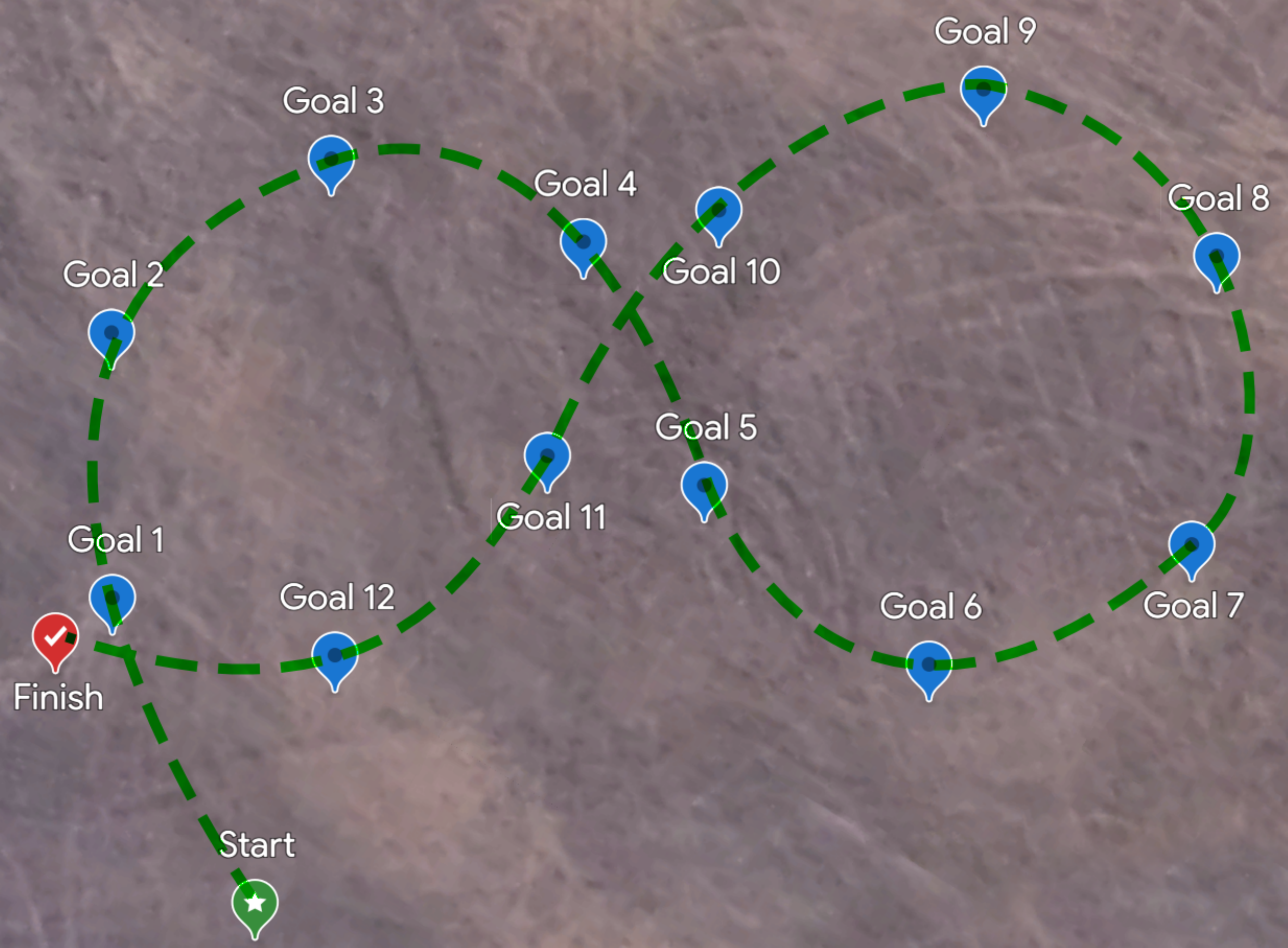}
    \caption{\textit{Figure-8 trajectory with waypoints spaced at 10m is used for real-world navigational experiments. Taking such sharp turns at a maximum speed of 5m/s tests the limits of all vehicle models}}
    \label{fig8}
    \vspace{-1em}
\end{figure}

We now demonstrate the impact of an accurate vehicle model on real-world navigation by navigating a challenging figure-8 trajectory (Fig. \ref{fig8}) at a max speed of 5m/s, allowing us to test PIAug's robustness to domain shift. A human first drives the desired maneuver to collect waypoints, as shown in Fig. \ref{fig8}.
Table \ref{tab: hardware} reports the results of our experiments. Note that while calculating the average velocity, we do not include the acceleration phase (Start to Goal 1) and deceleration phase (Goal 11 to Finish). Due to a long prediction horizon, the vehicle starts decelerating after it reaches Goal 11, as it can see the Finish point within its reach.

We can observe that when using a 4m goal radius, while PIAug and KBM both have a 100\% completion rate, the average velocity for PIAug is significantly higher than KBM. This happens since a high error in trajectory prediction, as is the case of KBM, leads the robot to correct its path often due to a significant difference between the actual state as compared to what was predicted in the past. This frequent correction can lead to disruptions when accelerating or maintaining high speed. Furthermore, the importance of a more accurate vehicle model is highlighted in the case of a 1m goal radius, where the experiment can only be completed in the presence of precise dynamics modeling. The average speed of PIAug increases when following a larger radius as the vehicle cuts a lot of corners and is not expected to follow the waypoints very precisely. This allows the vehicle to navigate the trajectory at a higher speed.

\begin{table}[t]
    \centering
    \vspace{0.3cm}
    \begin{tabular}{c|c||c|c}
        Goal Radius& Vehicle & No. of Successes & Mean Speed \\
        (m) & Model & (out of 3)&(m/s)\vspace{0.2em}\\
        \hline &&&\\[-1.5ex]
         4& KBM & 3 & 3.91 \vspace{0.2em}\\
         4&PIAug&3&\textbf{4.36} \vspace{0.2em}\\
         \hdashline &&&\\[-1.5ex]
         1& KBM & 0 & N/A \vspace{0.2em}\\
         1&PIAug&\textbf{3}&3.47 \vspace{0.2em}\\
    \end{tabular}
    \caption{\textit{ Effect of using KBM and PIAug on the overall Navigation Performance. N/A represents that the average speed cannot be calculated since no run was successful. 
    }}
    \vspace{-0.3cm}%
    \label{tab: hardware}
\end{table}
\section{Conclusion and Future Work}
\label{conclusion}
We introduce PIAug, a novel data augmentation method that uses a physics model to transfer data from the source to the target domain while integrating physics priors into NN training. We demonstrate its potential by developing a robust model for velocity distribution domain shift.
A comprehensive comparison of our model is done with other physics, data-driven, and hybrid models. Our approach outperforms the nominal model by reducing
 mean prediction error by upto 67\%. 
In real-life navigation experiments requiring precise trajectory tracking while traveling at out-of-domain velocities, our model is able to operate even under 4x tighter waypoint tracking constraints, surpassing KBM's limitations. These aggressive maneuvers highlight our methodology's robustness against domain-shifting. However, in contrast to low dimensional vehicle states, it is a challenging task to use physics priors to transfer environment modalities (e.g., FPV, terrain maps) to a sufficiently different target domain (e.g., transferring FPV data captured in a forest domain to an 
open area domain)

The future scope of this research is three-fold. First, physics-informed augmentation can reduce training time and data requirements for transferring a vehicle model across different robots using the target robot's nominal model.
Second, incorporating complex vehicle models \cite{magic_formula} can narrow the gap between augmented and real-world datasets. This is viable as the slower run times of these models only affect the training but not the inference time for PIAug-trained models.
Third, we can enhance safety by sampling augmentations from a task and constraint-informed prior distribution rather than random scaling.

\newpage

{
\bibliographystyle{IEEEtran}
\bibliography{refs}
}
\end{document}